\documentclass{article}
\usepackage{spconf,amsmath,epsfig}

\begin{document}\sloppy

% Example definitions.
% --------------------
\def\x{{\mathbf x}}
\def\L{{\cal L}}

% Title.
% ------
\title{ Mask-guided Style Transfer Network for Purifying Real Images }
%
% Single address.
% ---------------
\name{Tongtong Zhao, Yuxiao Yan , Jinjia Peng, Huibing Wang, Xianping Fu}
\address{Information Science and Technology College, Dalian Maritime University, Dalian 116026, China}

\maketitle

\begin{abstract}
Recently, the progress of learning-by-synthesis has proposed a training model for synthetic images, which can effectively reduce the cost of human and material resources. However, due to the different distribution of synthetic images compared with real images, the desired performance cannot be achieved. To solve this problem, the previous method learned a model to improve the realism of the synthetic images. Different from the previous methods, this paper try to purify real image by extracting discriminative and robust features to convert outdoor real images to indoor synthetic images. In this paper, we first introduce the segmentation masks to construct RGB-mask pairs as inputs, then we design a mask-guided style transfer network to learn style features separately from the attention and bkgd(background) regions and learn content features from full and attention region. Moreover, we propose a novel region-level task-guided loss to restrain the features learnt from style and content. Experiments were performed using mixed studies (qualitative and quantitative) methods to demonstrate the possibility of purifying real images in complex directions. We evaluate the proposed method on various public datasets, including LPW, COCO and MPIIGaze. Experimental results show that the proposed method is effective and achieves the state-of-the-art results.
\end{abstract}
\section{Introduction}

Recent appearance-based gaze estimation is performed under outdoor conditions by using an annotated large-scale real image training dataset. However, annotating training data sets requires a lot of manual labor. To solve this problem, a training model on a synthetic image is preferred because the annotations are automatically available. But this solution has a drawback, the distribution between the real image and the synthetic image is quite different. Traditionally, the solution is to use unmarked actual data to improve the authenticity of the synthetic image from the simulator, such as SimGANs\cite{Shrivastava2016learning}, these methods only learn the global features without considering local features. In the gaze estimation task, after realization with simGANs, the shape of the pupil or the edge of the pupil might by changed , the gaze estimation error will be increased due to the wrong pupil center location. Thus, these methods cannot be applied to outdoor (field) scenes due to its weak training time and adaptability to different situations in the field. In a different manner, we try to purify real image by extracting discriminative and robust features to convert outdoor real images to indoor synthetic images. Synthetic images is more regular and easy to learn, meanwhile, the annotations are automatically available.

To avoid changing the shape of the pupil or the edge of the pupil, we propose an mask-guided style transfer network to learn both local and global features. The way to handle local features is to obtain the attention region (pupil or iris) by segmentation. Fortunately, with the rapid development of deep learning \cite{wu2018deep,wu2018andwhere} based image segmentation methods including FCN\cite{FCN},SegNet\cite{segnet},U-net\cite{unet},Mask R-CNN\cite{mask}, we can obtain much better mask. To learn the style and content information from synthetic images, we first introduce the segmentation masks to construct RGB-mask pairs as inputs, then we design a mask-guided style transfer network to learn style features separately from the attention and bgkd(background) region , learn content features from full and attention region. For feature extraction, our work is most directly related to the work initiated by Gates et al.\cite{Gatys2015}. The feature map of the deep convolutional neural network with differentiated training is used to achieve the breakthrough performance of the transfer of painting style. We train a feed-forward feature extraction network for image transformation tasks. Our network aims to learn as much as possible on the premise of synthetic distribution, to minimize the loss of content transmission, and to solve the problem of insufficient spatial alignment information caused by the gram matrix. To achieve this goal, we propose a loss network with a novel task-guided loss, the attention region ,background region and full image region will be calculated in different task.

Our contributions are presented in three folds:

1. We took the first step to consider the attention region in style transfer task and propose an mask-guided style transfer network to purify the real image, making it similar to indoor conditions while retaining annotation information. Different with previous work in refining the synthetic images with global features, we purified the real images with local and global features.

2.Our network not only considers the RGB color channel, but uses the segmentation masks to construct RGB-mask pairs as inputs. We learned style features separately from the attention and bkgd(background) region and learned content features from full and attention region.

3. We proposed a hybrid research method (qualitative and quantitative) for experiments on two tasks. The results show that the proposed architecture significantly purified the real image compared with the baseline methods. Meanwhile, We achieve the state-of-the-art results on gaze estimation task.

 \begin{figure*}[!htbp]
\centering
 \begin{minipage}[]{1\textwidth}
    \centering
     \includegraphics[width = 0.6\textwidth,angle=0]{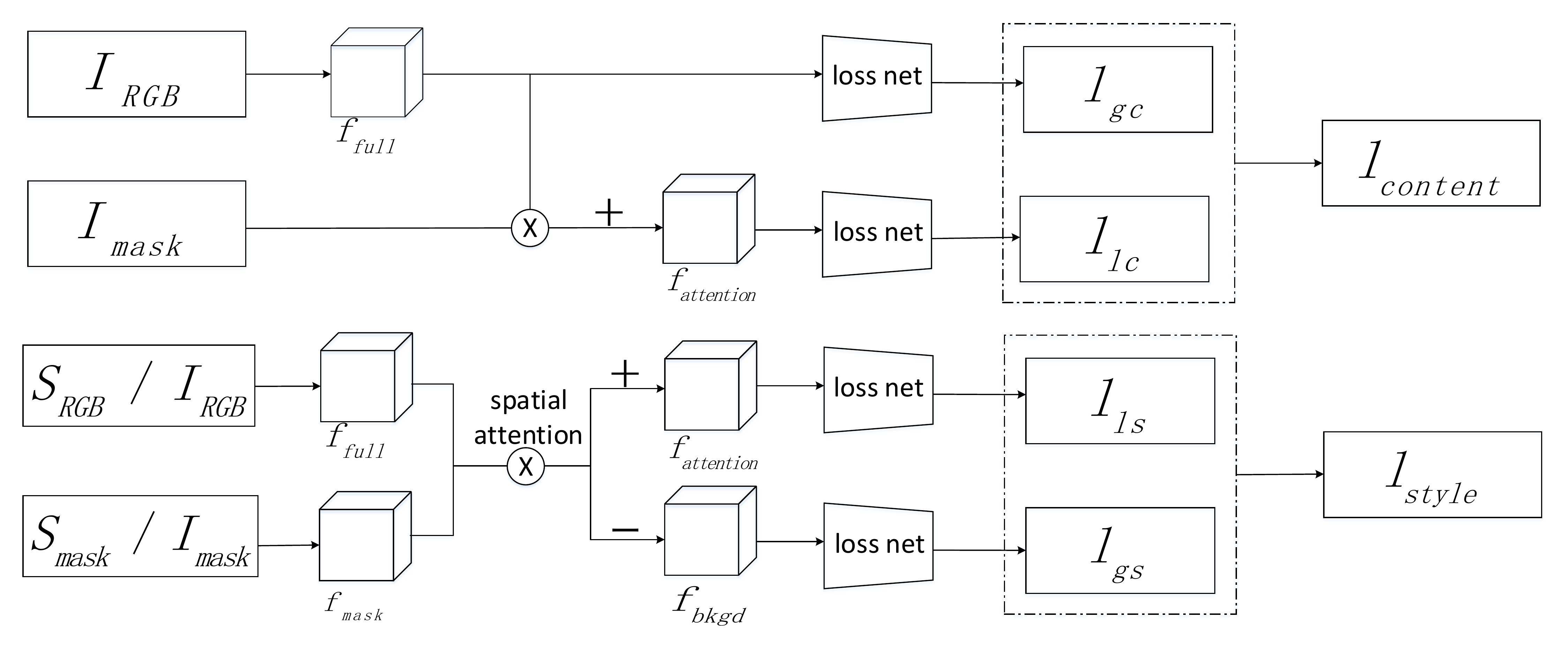}
 \end{minipage}
    \caption{Framework of proposed mask-guided style transfer network.}
    \label{figoverview}
\end{figure*}

\section{Proposed Method}\label{sec:Proposedmethod}

As shown in Figure 1, there are three multi-scale stages and a loss net to learn final features. It contains three multi-scale stages and a loss net to learn final features. There are three main streams which extracted from different regions of image , i.e. , the full-stream $f_{full}$, the attention stream $f_{attetion}$, the background stream $f_{bkgd}$. The full-stream $f_{full}$ learns features from the raw images. Meanwhile, the attention stream $f_{attetion}$ and the background stream $f_{bkgd}$ are learned attention features and background features with attention maps. The attention maps are generated by the attention subnet, the streams are designed to retain the content of input image $I_{RGB}$ and transfer the style from style image $S_{RGB}$ with the input image mask $I_{mask}$ and style image mask $S_{mask}$.

\subsection{Attention Subnet}

Given the input(style) image pair (RGB-Mask) as inputs, the attention subnet then produces attention maps which can be denoted as
\begin{equation}
att^{+}=\sigma(weight *(f_{full},f_{mask})+b )
\end{equation}
 where $\sigma$ is the sigmoid function, weight and $b$ are the convolutional filter weights and bias.  In the contrary, the background maps denoted as $att^{-}$ , $att^{+}$ and $att^{-}$ constitute a contrastive attention pair, for each location (i,j) which in the pair of attention maps and backgrounds maps should meet the constraint:
 \begin{equation}
 att^{+}(i,j)+att^{-}(i,j)=1
 \end{equation}
  Thus, the stream of attention and background can be denoted as :
\begin{equation}
\begin{split}
f_{attetion}=(f_{full},f_{mask})\otimes att^{+} \\
f_{bkgd}=(f_{full},f_{mask})\otimes att^{-}
\end{split}
\end{equation}
where $\otimes$ means the spatial weighting operation.

\subsection{Loss network with region-level task-guided loss }

With the attention maps described in last subsection, we further introduce the region-level triplet loss to enhance contrastive feature learning. After the attention operation, features from three main streams can be denoted as $f_{full}$,$f_{attetion}$ and $f_{bkgd}$, $f_{full}$,$f_{attetion}$ and $f_{bkgd}$ are used to calculate region-level task-guided loss for two tasks: keep the content and style transfer.

Our loss network can be divided into two parts: Feature reconstruction loss (a) and Style reconstruction loss(b), feature reconstruction loss is denoted as $\ell_{content}$ which is the summary of $\ell_{gc}$ and $\ell_{lc}$, meanwhile, style reconstruction loss is denoted as $\ell_{style}$ which is the summary of $\ell_{gs}$ and $\ell_{ls}$.

\subsubsection{Feature reconstruction loss}
Traditional feature reconstruction loss which known as content loss only takes the input image $I_{RGB}$ as input and try to minimize the loss between the content of input image $I_{RGB}$ and output image $O_{RGB}$ without considering encoding content reconstructions. We address this problem with the image segmentation masks $I_{mask}$ for the input images, the local feature of pupil region can be addressed when calculating the loss of $f_{full}$ and $f_{attetion}$. To visualise the image information that is encoded at different layers of the input image with masks, we perform gradient descent on a white noise image to find another image that matches the feature responses of the original image with mask. We then define the squared-error loss between the two feature representations
 \begin{equation}
\ell_{content}^l=\lambda_{g}\ell_{gc}^l+\lambda_{l}\ell_{lc}^l
\end{equation}
 \begin{equation}
\ell_{gc}^l=\sum_{c=1}^{C}\frac{1}{2N_{l}M_{l}}\sum_{ij}(F^{l,c}_{full}[O]-F^{l,c}_{full}[I])^{2}_{ij}
\end{equation}
 \begin{equation}
 \begin{split}
\ell_{lc}^l=\sum_{c=1}^{C}\frac{1}{2N_{l}M_{l}}\sum_{ij}(F^{l,c}_{full}[O]*F^{l,c}_{attetion}[I] \\
                              -F^{l,c}_{full}[I]*F^{l,c}_{attention}[I])^{2}_{ij}
\end{split}
\end{equation}

where  C is the number of channels in the semantic segmentation mask and $l$ indicates the $l$-th convolutional layer of the deep convolutional neural network, $F_{full}[\cdot]$ is the $f_{full}$ in each layer $l$ with the channel c, $F_{attention}[\cdot]$ is the $f_{attention}$ in each layer $l$ with the channel c, $\lambda_{g}$ is the weight to configure layer preferences of global losses ,$\lambda_{l}$ is the weight to configure layer preferences of local losses .

Each layer with $N_{l}$ distinct filters has $N_{l}$ feature maps each of size $M_{l}$, where $M_{l}$ is the height times the width of the feature map. So the responses in each layer $l$ can be stored in a matrix $F[\cdot] \in R^{N_{l}\times M_{l}}$ where $F[\cdot]_{ij}$ is the activation of the $i^{th}$ filter at position $j$ in each layer $l$.  As minimizing $\ell_{content}$, the image content and overall spatial structure are preserved but color, texture, and exact shape are not.

\subsubsection{Style reconstruction loss}

Feature Gram matrices are effective at representing texture, because they capture global statistics across the image due to spatial averaging. Since textures are static, averaging over positions is required and makes Gram matrices fully blind to the global arrangement of objects inside the reference image. So if we want to keep the global arrangement of objects, make the gram matrices more controllable to compute over the exact region of entire image, we need to add some texture information to the image.

Instead of taking input image $I_{RGB}$ and style image $S_{RGB}$ as inputs, we take the input image $I_{RGB}$ and style image $S_{RGB}$ with their mask $I_{mask}$ and $S_{mask}$ as pair inputs. To learn the skin style and pupil style respectively, we denote the pupil region as attention region and extract attention maps  $f_{attention}$ from both style image and input image, meanwhile, the skin region denoted as background region and product background maps $f_{bkgd}$ from style image and input image. We then define the squared-error loss between the two region feature representations
\begin{equation}
\ell_{style}^l=\lambda_{g}\ell_{gs}^l+\lambda_{l}\ell_{ls}^l
\end{equation}
 \begin{equation}
 \begin{split}
\ell_{gs}^l=\sum_{c=1}^{C}\frac{1}{4N^2_{l,c}M^2_{l,c}}\sum_{ij}\left(G^{l,c}_{bkgd}[O]-G^{l,c}_{bkgd}[S]\right)^2_{ij}
\end{split}
\end{equation}
\begin{equation}
\begin{split}
\ell_{ls}^l=\sum_{c=1}^{C}\frac{1}{4N^2_{l,c}M^2_{l,c}}\sum_{ij}\left(G^{l,c}_{attention}[O]-G^{l,c}_{attention}[S]\right)^2_{ij}
\end{split}
\end{equation}

where C is the number of channels in the semantic segmentation mask and $l$ indicates the $l$-th convolutional layer of the deep convolutional neural network. Each layer with $N_{l}$ distinct filters has $N_{l}$ feature maps each of size $M_{l}$, where $M_{l}$ is the height times the width of the feature map. So the responses in each layer $l$ can be stored in a matrix $F[\cdot] \in R^{N_{l}\times M_{l}}$ where $F[\cdot]_{ij}$ is the activation of the $i^{th}$ filter at position $j$ in each layer $l$. $G^{l,c}_{\cdot}[\cdot]$can be denoted as follows:
\begin{equation}
 \begin{split}
G^{l,c}_{bkgd}[O]=(F^{l,c}_{full}[O]*F^{l,c}_{bkgd}[I]) \\
                  *(F^{l,c}_{full}[O]*F^{l,c}_{bkgd}[I])^{T}
\end{split}
\end{equation}
\begin{equation}
 \begin{split}
G^{l,c}_{bkgd}[S]=(F^{l,c}_{full}[S]*F^{l,c}_{bkgd}[S]) \\
                  *(F^{l,c}_{full}[S]*F^{l,c}_{bkgd}[S])^{T}
\end{split}
\end{equation}
\begin{equation}
 \begin{split}
G^{l,c}_{attention}[O]=(F^{l,c}_{full}[O]*F^{l,c}_{attention}[I]) \\
                       *(F^{l,c}_{full}[O]*F^{l,c}_{attention}[I])^{T}
\end{split}
\end{equation}
\begin{equation}
 \begin{split}
G^{l,c}_{bkgd}[S]=(F^{l,c}_{full}[S]*F^{l,c}_{attention}[S]) \\
                 *(F^{l,c}_{full}[S]*F^{l,c}_{attention}[S])^{T}
\end{split}
\end{equation}
where $F_{full}[\cdot]$ is the $f_{full}$ in each layer $l$ with the channel c, $F_{attention}[\cdot]$ is the $f_{attention}$ in each layer $l$ with the channel c, $F_{bkgd}[\cdot]$ is the $f_{bkgd}$ in each layer $l$ with the channel c, $\lambda_{g}$ is the weight to configure layer preferences of global losses ,$\lambda_{l}$ is the weight to configure layer preferences of local losses.

We formulate the style transfer objective by combining both two components together:
\begin{equation}
L_{total}=\sum_{l=1}^L\alpha_l\ell_{content}^l+\sum_{l=1}^L\beta_l\ell_{style}^l
\end{equation}
where L is the total number of convolutional layers and $l$ indicates the $l$-th convolutional layer of the deep convolutional neural network. $\alpha_{l}$ and $\beta_{l}$ are the weights to configure layer preferences. $\ell_{content}$ is the content loss (Eq.(10)) and $\ell_{style}$ is the style loss(Eq.(13)). $\alpha_l$,$\beta_l$ are scalars, $\alpha_l=10^{2}$,$\beta_l=10^{4}$, in all cases the hyperparameters $\alpha_l$,$\beta_l$ are exactly the same.  We find that unconstrained optimization of Equation 20 typically results in images whose pixels fall outside the range [0,255]. For a more fair comparison with our method whose output is constrained to this range, for the baseline we minimize Equation 20 using projected L-BFGS. Image O is generated by solving the problem
 \begin{equation}
O=\arg\min_{I}{L_{total}}+\theta\ell_{TV}(I)
\end{equation}
where I is initialized with white noise. The advantage of this solution is that the requirement for mask is not too precise. It does not only retain the desired structural features, but also enhance the estimation of the pupil and iris information during the reconstruction of the style.

\section{Experimental Results}\label{sec:experimentalresults}

\subsection{Style Transfer}

The purpose of the style transfer is to generate an image that combines the content of the target content image as the real image content with the style of the target style image as the style of the synthetic image. We train an image transformation network for each of the several hand selection style goals and compare our results with the baseline methods of Gatys et al.\cite{Gatys2015} and Feifei Li et al.\cite{lifeifei2016}. As a baseline, we re-implemented the method of Gatys et al.\cite{Gatys2015} and Feifei Li et al.\cite{lifeifei2016}. In order to make a fairer comparison with our method whose output is constrained to [0, 255], for the baseline, we minimize the equation 1 and equation 6 by using the projected L-BFGS by cropping the image to the range [0, 255] at each iteration. In most cases, the optimization converges to satisfactory results in 500 iterations.

\begin{figure}[!htbp]
\centering
 \begin{minipage}[]{0.5\textwidth}
    \centering
     \includegraphics[width = 1\textwidth,angle=0]{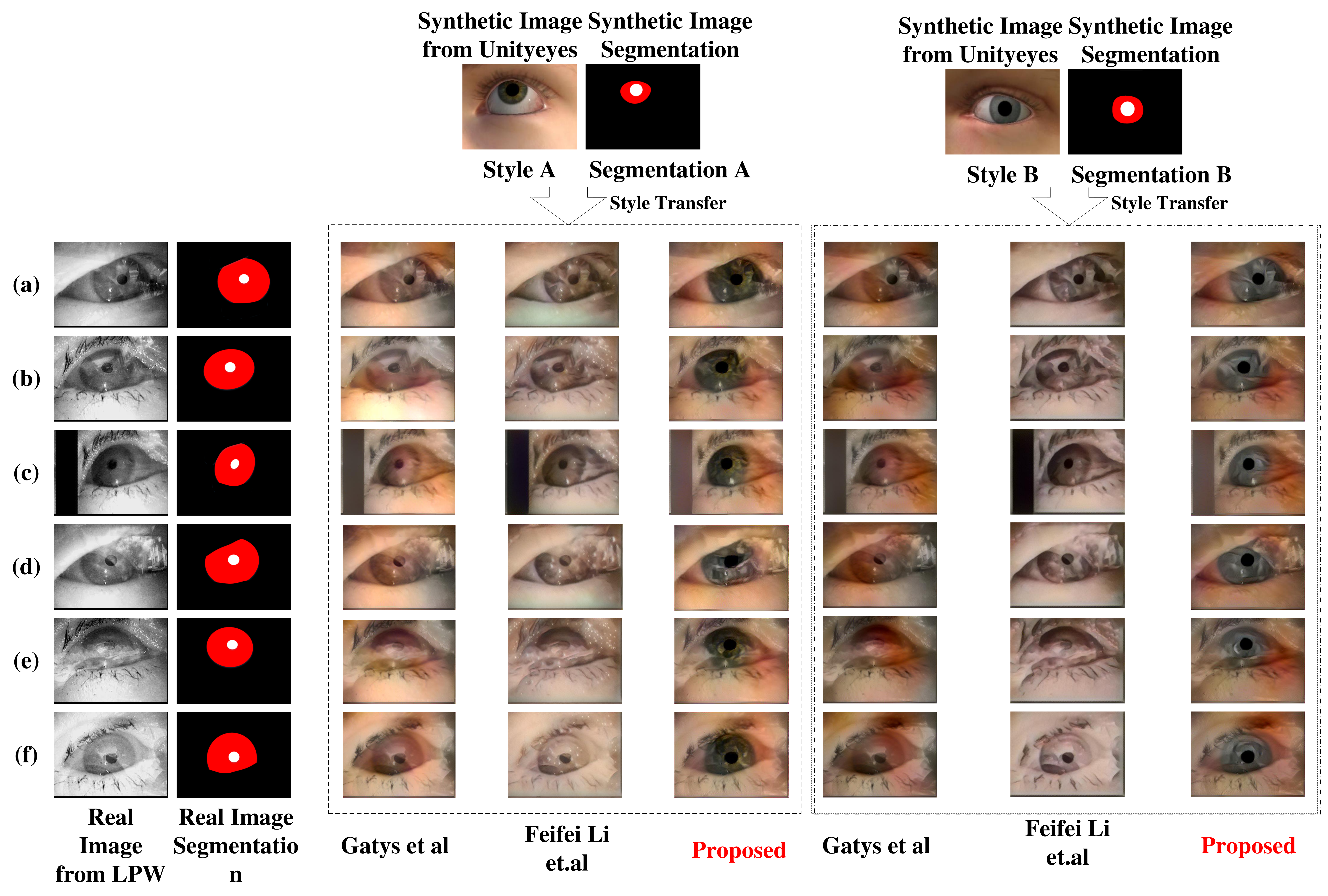}
 \end{minipage}
    \caption{Comparison on public LPW dataset with available style transfer methods.(a),(b),(c),(d),(e),and (f) represent the purified results of different distributions under six outdoor conditions from LPW dataset with three different styles from UnityEyes dataset. Style A, B represent two different distributions of indoor conditions. The distribution of pupil and iris regions is dramatically different from style image. The proposed method, therefore can separate the pupil and the iris regions easily and the distribution of pupil and iris regions is similar to style image. }
    \label{figstyle}
\end{figure}

$\mathbf{Qualitative}$ $\mathbf{Results}$:
We compare the proposed style transfer method with methods proposed by Gatys et al.\cite{Gatys2015} and Feifei Li et al.\cite{lifeifei2016}. Figure 4 shows the qualitative results in the indoor and outdoor scenes for the UnityEyes\cite{Wood2016learning} and LPW\cite{LPW2016} datasets respectively. From the LPW dataset, we choose six different real images to represent six different conditions of outdoor scenes, which are images (a),(b),(c),(d),(e), and (f). From the UnityEyes dataset, we selected two synthetic images that have different distributions as the style images, style A, B, none of which has no same gaze angle as the six real images. As can be seen from (a),(b), and (c), the proposed method is less affected by external factors, such as illumination, and is similar to the results obtained by Gatys et al. \cite{Gatys2015} and Feifei Li et al.\cite{lifeifei2016}. However, if look closely, we can find that the proposed method is more capable of preserving the color information of the style image. From (d),(e), and (f), it is obvious that Gatys et al.\cite{Gatys2015} and Feifei Li et al.\cite{lifeifei2016} are so greatly influenced by light and other factors that the pupil and the iris regions cannot be completely separated and even in (e),there is a loss of pupil area. What's more, the distribution of pupil and iris regions is dramatically different from style image. In comparison, our proposed method can not only separate the pupil and the iris regions more easily but also the distribution of pupil and iris regions is more similar to style image.

\begin{table}[!hbp]
\centering
\label{tabLr1}
\caption{ Speed (in seconds) for our style transfer networks vs Gatys et al. \cite{Gatys2015}, Feifei Li et al.\cite{lifeifei2016} for various resolutions. All benchmarks use a Titan X GPU.}
\begin{tabular}{|l|c|c|c|}
\hline
Method& $256\times256$ & $512\times512$ & $1024\times1024$\\
\hline
[1]Gatys&12.69s & 45.88s & 171.55s\\
\hline
[2]Feifei li& 0.023s & 0.082s & 0.351s \\
\hline
[3]Our Method & $\mathbf{0.015s}$ &$\mathbf{0.048s}$ & $\mathbf{0.211s}$ \\
\hline
[3] vs [1] speedup & $\mathbf{1060x}$ & $\mathbf{1026x}$ & $\mathbf{1042x}$ \\
\hline
[3] vs [2] speedup & $\mathbf{1.53x}$ & $\mathbf{1.6x}$ & $\mathbf{1.67x}$ \\
\hline
\end{tabular}
\end{table}

$\mathbf{Speed}$: Table 2 compares the runtime of our method with Gatys et al.\cite{Gatys2015}, Feifei Li et al.\cite{lifeifei2016} for several image sizes. Across all image sizes, compared to 400 iterations of the baseline methods, our method is three orders of magnitude faster than Gatys et al.\cite{Gatys2015} and we achieve better qualitative results (Fig.6) compared with Feifei Li et al.\cite{lifeifei2016} in tolerate speed. Our method processes images of size $512\times512$  at 20 FPS, making it feasible to run in real-time or on video.

\subsection{Appearance-based Gaze Estimation}

$\mathbf{Implementation }$ $\mathbf{Details}$: In order to verify the effectiveness of the proposed method for gaze estimation, 3 public datasets (UTView\cite{Sugano2014learning}, SynthesEyes\cite{Wood2015rendering}, UnityEyes\cite{Wood2016learning}) are used to train the estimator with k-NN\cite{ALearning}\cite{AMultiview}\cite{ARobust}\cite{ASemantic}and CNN\cite{ACycle-Consistent,ADeep,wu20193D,AEffective,AIterative}. MPIIGaze dataset\cite{MPIIGaze2017} and purified MPIIGaze dataset (purified by proposed method) are used for testing the accuracy.

\begin{table}[!hbp]
\centering
\label{tabLr1}
\caption{Test performance on MPIIGaze and purified MPIIGaze; Purified MPIIGaze is the dataset which purified by proposed method. "Method" represents training set used with gaze estimation method.  Note how purifying real dataset for training lead to improved performance. }
\begin{tabular}{|l|c|c|}
\hline
Method & MPIIGaze & purified MPIIGaze \\
\hline
SVR &16.5$^{\circ}$ & 14.3$^{\circ}$ \\
\hline
ALR &16.4$^{\circ}$ & 13.9$^{\circ}$ \\
\hline
Random Forest(RF) & 15.4$^{\circ}$ & 14.2$^{\circ}$ \\
\hline
KNN with UTview &16.2$^{\circ}$ & 13.6$^{\circ}$ \\
\hline
CNN with UTview &13.9$^{\circ}$ & 11.7$^{\circ}$ \\
\hline
KNN with UnityEyes  &12.5$^{\circ}$ & 9.9$^{\circ}$ \\
\hline
CNN with UnityEyes &11.2$^{\circ}$ & 7.8$^{\circ}$ \\
\hline
KNN with Syntheyes  &11.4$^{\circ}$ &8.0$^{\circ}$  \\
\hline
CNN with Syntheyes  &13.5$^{\circ}$ & 8.8$^{\circ}$ \\
\hline
\end{tabular}
\end{table}

$\mathbf{Quantitative }$ $\mathbf{Results}$:
There are five gaze estimation methods used as base-line estimation methods. Three of them are common methods, Support Vector Regression(SVR), Adaptive Linear Regression(ALR) and Random Forest(RF).
The other two methods are reproduced for fairly comparison with state-of-the-art. The first method is a simple cascaded method\cite{wang2017}\cite{ACML2018}\cite{ICIMCS2017} which uses multiple $k$-NN($k$-Nearest Neighbor) classifier to select neighbors in feature space joint head pose, pupil center and eye appearance. The second one aims to train a simple convolutional neural network (CNN)\cite{wild2017}\cite{IJCNN2018}\cite{PR2018} to predict the eye gaze direction by using $ l_{2}$ loss. As is shown in table 2, we compare the performance of these two gaze estimation methods on different datasets, where the "Method" represents gaze estimation methods with different training sets. It can be observed that the accuracy of gaze estimation of each dataset has improved at least three degrees, which means that our proposed method has greatly improved the performance when testing the output. This improvement will have some practical value in human-computer interaction.

\begin{table}[!hbp]
\centering
\label{tabLr1}
\caption{Test performance on raw MPIIGaze(R MPIIGaze) and purified MPIIGaze(P MPIIGaze); Purified MPIIGaze is the dataset which purified by proposed method. "Method" represents training set used with gaze estimation method.  Note instead of refining training synthetic dataset, purifying testing real dataset is a better choice for gaze estimation. }
\begin{tabular}{|l|c|c|}
\hline
Method & R/P  MPIIGaze & error \\
\hline
CNN with UnityEyes & R MPIIGaze &11.2 \\
\hline
CNN with Refined UnityEyes& R MPIIGaze & 9.9 \\
\hline
CNN with UnityEyes & P MPIIGaze & 7.8 \\
\hline
\end{tabular}
\end{table}

Furthermore, in order to prove that purifying real dataset can achieve better performance than refining synthetic dataset on gaze estimation task, we compare proposed method with state-of-the-art methods of refining synthetic dataset for gaze estimation,SimGANs\cite{Shrivastava2016learning}. In order to make a fairer comparison with our method, we reproduce SimGANs\cite{Shrivastava2016learning}, different from traditional methods which training on synthetic UnityEyes dataset and testing on real MPIIGaze dataset, SimGANs\cite{Shrivastava2016learning} refining UnityEyes with its method, try to realistic synthetic image and test on real MPIIGaze dataset. In the contrary, proposed method purify the testing real MPIIGaze dataset without modifying training synthetic datset. From Table 3, we can see that purifying testing real dataset outperforms traditional methods and SimGans\cite{Shrivastava2016learning} on gaze estimation.

$\mathbf{Preserving }$ $\mathbf{Ground}$ $\mathbf{Truth}$: We manually quantify the gaze center of the 200 real and purified images by fitting the ellipse to the pupil, which can be seen as an approximation of the direction of gaze and is difficult for humans to label accurately, to quantify that there is no significant change in the ground truth gaze direction. The absolute difference of the estimated pupil center between the real and corresponding purified images is very small: 0.8 $\pm$ 1.1 (eye width=55px)
\begin{figure}[!htbp]
\centering
 \begin{minipage}[]{0.5\textwidth}
    \centering
     \includegraphics[width = 0.8\textwidth,angle=0]{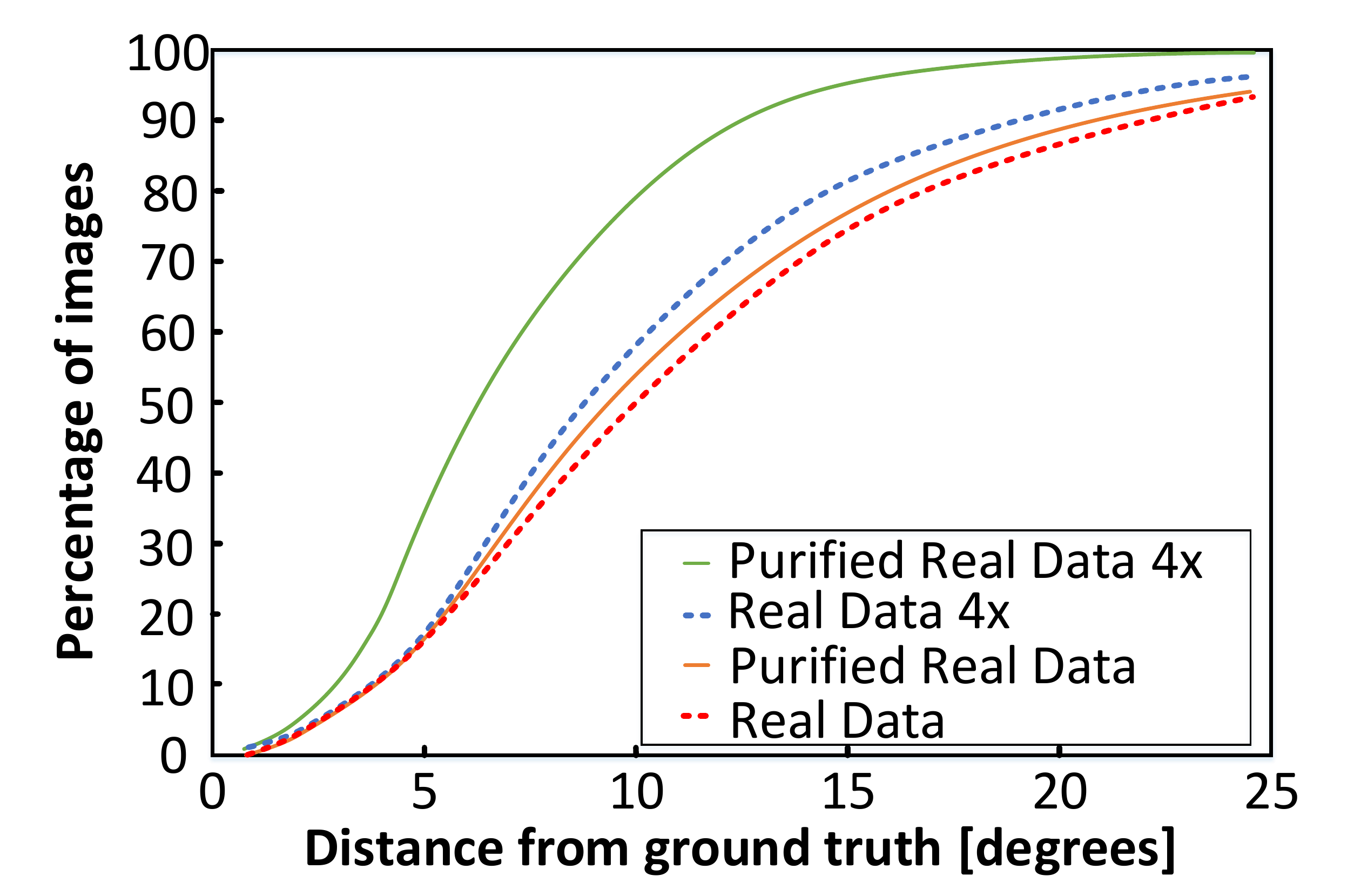}
 \end{minipage}
    \caption{Quantitative results for appearance-based gaze estimation on the MPIIGaze dataset and purified MPIIGaze dataset. The plot shows cumulative curves as a function of degree error as compare to the ground truth eye gaze direction, for different numbers of testing examples of data.}
    \label{figstyle}
\end{figure}

\section{Conclusion}\label{sec:conclusion}

This paper purify the real image by weakening its distribution, which is a better choice than improving the realism of synthetic image. We have applied this method to style transfer and gaze estimation tasks where we achieved comparable performance and drastically improved speed compared with existing methods. Performance evaluation indicates that purified MPIIGaze dataset (purified by our proposed method) records smaller error angle when used for gaze estimation task as compared with the raw MPIIGaze dataset.

In the future, we intend to explore modeling the real-time gaze estimation system based on the proposed method and improve the speed of purifying videos.

% References should be produced using the bibtex program from suitable
% BiBTeX files (here: strings, refs, manuals). The IEEEbib.bst bibliography
% style file from IEEE produces unsorted bibliography list.
% -------------------------------------------------------------------------
\bibliographystyle{IEEEbib}
\bibliography{icme2019template}

\end{document}